\documentclass{article}

\usepackage{amsmath,amsfonts,amssymb,times,graphicx,natbib,algorithm,algorithmic,hyperref}

\usepackage[accepted]{data4good2016}

\icmltitlerunning{Quantifying and Reducing Stereotypes in Word Embeddings}

\usepackage{comment}
\usepackage{adjustbox}
\usepackage{mathtools}

\newcommand{\ignore}[1]{}

\begin{document}

\twocolumn[
\icmltitle{Quantifying and Reducing Stereotypes in Word Embeddings}
\vspace{-0.4cm}
\icmlauthor{Tolga Bolukbasi\texorpdfstring{\textsuperscript{1}}{}}{tolgab@bu.edu}
\icmlauthor{Kai-Wei Chang\texorpdfstring{\textsuperscript{2}}{}}{kw@kwchang.net}
\icmlauthor{James Zou\texorpdfstring{\textsuperscript{2}}{}}{jamesyzou@gmail.com}
\icmlauthor{Venkatesh Saligrama\texorpdfstring{\textsuperscript{1}}{}}{srv@bu.edu}
\icmlauthor{Adam Kalai\texorpdfstring{\textsuperscript{2}}{}}{adam.kalai@microsoft.com}
\icmladdress{1 Boston University, 8 Saint Mary's Street, Boston, MA}
\vspace{-0.3cm}
\icmladdress{2 Microsoft Research New England, 1 Memorial Drive, Cambridge, MA}
\vskip 0.1cm
]

\begin{abstract}
Machine learning algorithms are optimized to model statistical properties of the training data. If the input data reflects stereotypes and biases of the broader society, then the output of the learning algorithm also captures these stereotypes.  In this paper, we initiate the study of gender stereotypes in {\em word embedding}, a popular framework to represent text data. As their use becomes increasingly common, applications can inadvertently amplify unwanted stereotypes. We show across multiple datasets that the embeddings contain significant gender stereotypes, especially with regard to professions. We created a novel gender analogy task and combined it with crowdsourcing to systematically quantify the gender bias in a given embedding. We developed an efficient algorithm that reduces gender stereotype using just a handful of training examples while preserving the useful geometric properties of the embedding. We evaluated our algorithm on several metrics. While we focus on male/female stereotypes, our framework may be applicable to other types of embedding biases.
\vspace{-0.4cm}
\end{abstract}
\section{Introduction}

Word embeddings, trained only on word co-occurrence in text corpora, capture rich semantic information about words and their 
meanings~\cite{mikolov2013linguistic}. Each word (or common phrase) $w \in W$ is 
encoded as a $d$-dimensional {\em word vector} $v_w \in \mathbb{R}^d$. Using 
simple vector arithmetic, the embeddings are capable of answering 
{\em analogy puzzles}. For instance, {\em man}:{\em king} :: {\em 
woman}:\textit{\underline{\ \ \ \ }}\footnote{
An {\em analogy 
puzzle}, $a$:$b$ :: $c$:$d$, involves selecting the most appropriate $d$ given $a$, $b$, and $c$.} 
returns {\em queen} as the answer, and similarly Japan is returned for {\em Paris}:{\em France} :: {\em 
Tokyo}:\underline{\em Japan} (computer-generated answers are underlined). 
A number of such embeddings have been made publicly available including the popular word2vec~\cite{MCCD13,MSCCD13} embedding trained on 3 million words into 300 
dimensions, which we refer to here as the w2vNEWS embedding because it was trained on a corpus of text from Google News. These word embeddings have been used in a variety of downstream applications (e.g., document ranking~\cite{NMCC16}, sentiment analysis~\cite{IrsoyCardie14}, and question retrieval~\cite{LJBJTMM16}).  

While word-embeddings encode semantic information they also exhibit hidden biases inherent in the dataset they are trained on. 
For instance, word embeddings based on w2vNEWS can return biased solutions to analogy puzzles such as {\em father}:{\em doctor} :: {\em mother}:\textit{\underline{nurse}} and {\em man}:{\em computer programmer} :: {\em woman}:\textit{\underline{homemaker}}.  Other publicly available embeddings produce similar results exhibiting gender stereotypes. Moreover, the closest word to the query {\em BLACK MALE} returns {\em ASSAULTED} while the response to {\em WHITE MALE} is {\em ENTITLED TO}. This raises serious concerns about their widespread use. 

The prejudices and stereotypes in these embeddings reflect biases implicit in the data on which they were trained. The embedding of a word is typically optimized to predict co-occuring words in the corpus. Therefore, if {\em mother} and {\em nurse} frequently co-occur, then the vectors $v_{mother}$ and $v_{nurse}$ also tend to be more similar and encode the gender stereotypes. The use of embeddings in applications can amplify these biases. To illustrate this point, consider Web search where, for example, one recent project has shown that, when carefully combined with existing approaches, word vectors can significantly improve Web page relevance results \cite{NMCC16} (note that this work is a proof of concept -- we do not know which, if any, mainstream search engines presently incorporate word embeddings). Consider a researcher seeking a summer intern to work on a machine learning project on deep learning who searches for, say, ``linkedin graduate student machine learning neural networks.'' Now, a word embedding's semantic knowledge can improve relevance in the sense that a LinkedIn web page containing terms such as ``PhD student,'' ``embeddings,'' and ``deep learning,'' which are related to but different from the query terms, may be ranked highly in the results. However, word embeddings also rank CS research related terms closer to male names than female names. The consequence would be, between two pages that differed in the names Mary and John but were otherwise identical, the search engine would rank John's higher than Mary. In this hypothetical example, the usage of word embedding makes it even harder for women to be recognized as computer scientists and would contribute to widening the existing gender gap in computer science. While we focus on gender bias, specifically male/female, our approach may be applied to other types of biases.  

We propose two methods to systematically quantify the gender bias in a set of word embeddings. First, we quantify how words, such as those corresponding to professions, are distributed along the direction between embeddings of {\em he} and {\em she}. Second, we design an algorithm for generating analogy pairs from an embedding given two seed words and we use crowdworkers to quantify whether these embedding analogies reflect stereotypes. Some analogies reflect stereotypes such as {\em he}:\underline{\em janitor} :: {\em she}:\underline{\em housekeeper} and {\em he}:\underline{{\em alcoholism}} :: {\em she}:\underline{{\em eating disorders}}. Finally, others may provoke interesting discussions such as {\em he}:\underline{\em realist} :: {\em she}:\underline{\em feminist} and {\em he}:\underline{\em injured} :: {\em she}:\underline{\em victim}.
 

Since biases are cultural, we enlist U.S.-based crowdworkers to identify analogies to judge whether analogies: (a) reflect {\em stereotypes} (to understand biases), or (b) are nonsensical (to ensure accuracy). We first establish that biases indeed exist in the embeddings. We then show that, surprisingly, information to distinguish stereotypical associations like female:homemaker from definitional associations like female:sister can often be removed. We propose an approach that, given an embedding and only a handful of words, can reduce the amount of bias present in that embedding without significantly reducing its performance on other benchmarks.


\textbf{Contributions.} 
%
\textbf{(1)} We initiate the study of stereotypes and biases in word embeddings. Our work follows a large body of literature on bias in language, but word embeddings are of specific interest because they are commonly used in machine learning and they have simple geometric structures that can be quantified mathematically. 
%
\textbf{(2)}
We develop two metrics to quantify gender stereotypes in word embeddings based on words associated with professions together with automatically generated analogies which are then scored by the crowd. 
%
\textbf{(3)}
We develop a new algorithm that reduces gender stereotypes in the embedding using only a handful of training examples while preserving useful properties of the embedding. 

\begin{figure}
\includegraphics[width=\linewidth]{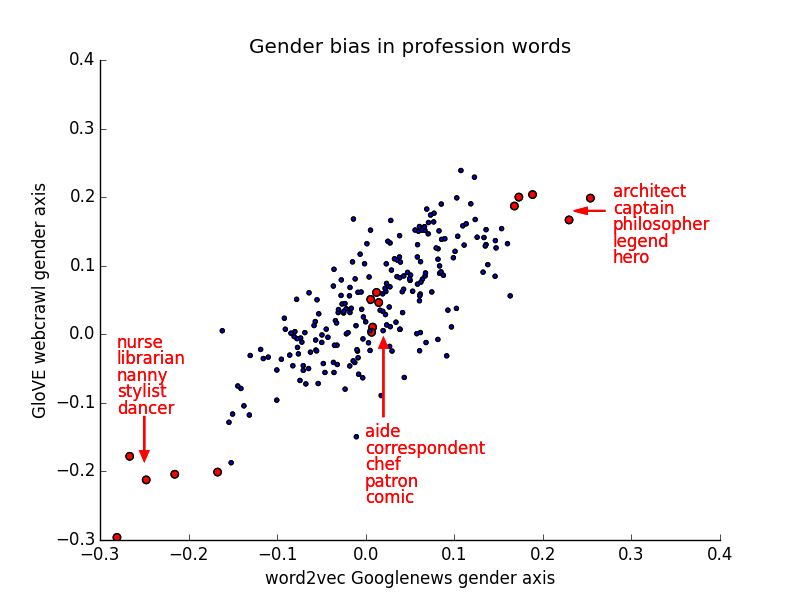}
\caption{Comparison of gender bias of profession words across two embeddings: word2vec trained on Googlenews and GloVe trained web-crawl texts. The $x$ and $y$ axes show projections onto the \emph{he-she} direction in the two embeddings. Each dot is one of 249 common profession words. Words closest to \emph{he}, closest to \emph{she}, and in between the two are colored in red and shown in the plot.}
\label{fig:crosscorpus}
\vspace{-0.6cm}
\end{figure}

{\bf Prior work.} The body of prior work on bias in language and prejudice in machine learning algorithms is too large to fully cover here. We note that gender stereotypes have been shown to develop in children as young as two years old \cite{turner1995multidimensional}. Statistical analyses of language have shown interesting contrasts between language used to describe men and women, e.g., in recommendation letters \cite{schmader2007linguistic}. A number of online systems have been shown to exhibit various biases, such as racial discrimination in the ads presented to users \cite{sweeney2013discrimination}. Approaches to modify classification algorithms to define and achieve various notions of fairness have been described in a number of works, see, e.g., \cite{barocas2014big, dwork2012fairness} and a recent survey \cite{zliobaite2015survey}.

\section{Implicit stereotypes in word embedding}

\textbf{Stereotyped words}. A simple approach to explore how gender stereotypes manifest in embeddings is to quantify which words are closer to {\em he} versus {\em she} in the embedding space (using other words to capture gender, such as {\em man} and {\em woman}, gives similar but noisier results due to their multiple meanings). We used a list of 215 common profession names, removing names that are associated with one gender by definition (e.g. waitress, waiter). For each name, $v$, we computed its projection onto the gender axis: $v\cdot (v_{he} - v_{she})/||v_{he} - v_{she}||_2$. Figure~1 shows the projection of professions on the w2vNEWS embedding ($x$-axis) and on a different embedding trained by GloVe on a dataset of web-crawled texts ($y$-axis). Several professions are closer to the {\em he} or {\em she} vector and this is consistent across the embeddings, suggesting that embeddings encode gender stereotypes. 

\textbf{Stereotyped analogies}. While professions give easily-interpretable insights on embedding stereotypes, we developed a more general method to automatically detect and quantify gender bias in any word embedding. Embeddings have shown to perform well in analogy tasks. Motivated by this, 
 we ask the embedding to generate analogous word pairs for {\em he} and {\em she}, and use crowd-sourcing to evaluate the degree of stereotype of each pair. 


A desired analogy {\em he}:{\em she} :: $w_1$:$w_2$ has the following properties\footnote{For the ease of presentation, we abuse the notation to use $w$ to represent a word or a word vector depending on the context.}: 1) the direction of $w_1$-$w_2$ has to align with {\em he}-{\em she}; 2) $w_1$ and $w_2$ should be semantically similar, i.e. $||w_1-w_2||_2$ is not too large.
Based on this, given a word embedding $E$, we proposed to score analogous pairs by the following formulation:
\begin{equation}
\label{eq:analogyS}
\mathclap{\mbox{S}_d(w_a, w_b) =  \frac{(w_a-w_b)\cdot d}{||w_a - w_b||_2} \mbox{ s.t. } ||w_a - w_b||_2 \leq \delta}
\end{equation}
where $d = (v_{he} - v_{she})/||v_{he} - v_{she}||_2$ is the gender direction and $\delta$ is a threshold for similarity.\footnote{We explored alternatives including a variation of 
3-CosMul~\cite{levyGoldberg14} for generating word pairs, and observe that the proposed 
approach works the best.} 
 We observe that setting $\delta =1$ often works well in practice; this corresponds to requiring that the two words forming the analogy are significantly closer together than two random embedding vectors.  
 
From the embedding, we generated the top analogous pairs with the largest $S_d$ scores. To avoid redundancies, if multiple pairs share the same $w_a$ or $w_b$, we kept only one pair. Then we employed Amazon Mechanical Turk to evaluate the analogies. For each analogy, such as {\em man}:{\em woman} :: {\em doctor}:{\em nurse}, we ask the Turkers 
two yes/no questions to verify if this pairing makes sense as an analogy and whether it exhibits gender stereotype.
Every word pair is judged by 10 Turkers, and we used the number of Turkers that rated this pair as stereotyped to quantify the degree of bias of this analogy.  
Table \ref{fig:most-stereotypical} shows the most and least stereotypical 
analogies generated by word2vec on Googlenews. 
Overall, 21\% and 32\% analogy judgments were 
stereotypical and nonsensical, respectively, by the Turkers. 

\begin{table*}
\begin{tabular}{cccc}
\multicolumn{4}{c}{\bf{Ranked as M-F stereotypical by 10/10 workers:} }\\ 
{\em surgeon:nurse} & {\em doctors:midwives} & {\em athletes:gymnasts} & {\em paramedic:registered nurse}\\
{\em Hummer:minivan} & {\em Karate:Gymnastics} & {\em woodworking:quilting} & {\em alcoholism:eating disorders}\\
{\em athlete:gymnast} & {\em neurologist:therapist} & {\em hockey:figure skating} & {\em architect:interior designer}\\
{\em chauffeur:nanny} & {\em curator:librarian} & {\em pilots:flight attendant} & {\em drug trafficking:prostitution}\\
{\em musician:dancer} & {\em beers:cocktails} & {\em Sopranos:Real Housewives} & {\em headmaster:guidance counselor}\\
{\em workout:Pilates} & {\em Home Depot:JC Penney} & {\em weightlifting:gymnastics} & {\em Sports Illustrated:Vanity Fair}\\
{\em carpentry:sewing} & {\em accountant:paralegal} & {\em addiction:eating disorder} & {\em professor emeritus:associate professor}\\
\multicolumn{4}{c}{\bf{Ranked as M-F stereotypical by 0/10 workers:} (random sample of 12 out of 505)}\\
{\em Jon:Heidi} & {\em Ainge:Fulmer} & {\em Allan:Lorna} & {\em George Clooney:Penelope Cruz}\\
{\em Erick:Karla} & {\em gentlemen:ladies} & {\em Christopher:Jennifer} & {\em veterans:servicemen}\\
{\em sausages:buns} & {\em patriarch:matriarch} & {\em Leroy:Lucille} & {\em Phillip:Belinda}\\\end{tabular}
\caption{\label{fig:most-stereotypical}Sample of the top 1,000 analogies generated for {\em he}:{\em she} :: $w_a$:$w_b$ on w2vNEWS, ordered by the number 
of workers who judged them to reflect stereotypes. The analogies which were rated stereotypical by 10/10 workers are shown and a random sample of twelve analogies 
rated as stereotypical by 0/10 workers is shown. Overall, 21\% of the 1000 analogies were rated as reflecting gender stereotypes.}
\vspace{-0.3cm}
\end{table*}
\vspace{-0.3cm}
\section{Reducing stereotypes in word embedding}
\label{sec:debias}

Having demonstrated that word embeddings contain substantial stereotypes in both professions and analogies, we developed a method to reduce these stereotypes while preserving the desirable geometry of the embedding. 

Word embeddings are often trained on a large corpus (w2vNEWS is trained on 
Google news corpus with 100 billion words). As a result, it is impractical and even 
impossible (the corpus is not publicly accessible) to reduce the stereotypes during the training of the the word 
vectors. Therefore, we assume that we are given a a set of word vectors and aim to remove stereotypes as a post-processing step.   
   
Our approach takes the following as inputs:
\textbf{(1)} a word embedding stored in a matrix $E \in \mathbb{R}^{n,r}$, where $n$ is the number of words and $r$ 
is the dimension of the latent space. \textbf{(2)} A matrix $B \in \mathbb{R}^{n_b, r}$ where each column 
is a vector representing a direction of stereotype. In this paper, $B = v_{he} - v_{she}$, but in general, $B$ can contain multiple stereotypes including 
gender, racism, etc. \footnote{Here we assume the stereotypical 
directions are given. In practice, this can be obtained by subjecting the
vectors of the extreme words in the concept (e.g. {\em he} and {\em she} representing gender.)}
\textbf{(3)} A matrix $P \in \mathbb{R}^{n_p, r}$ whose columns correspond to set of seed words that we want to debias. An example of a seed word for gender is {\em manager}.  
\textbf{(4)} A matrix $A\subseteq E$ whose columns represent a background set of words. We want the algorithm  
to preserve their  pairwise distances.\footnote{
Typically, we can set $A$ to contain the word vectors in $E$ except the ones
in $B$ and $P$.} 

The goal is to  
generate a transformation matrix $\hat{T} \in N^{r,r}$, 
which has the following properties:\\
\textbullet $\:$ The transformed  embeddings are stereotypical-free.   That is  every column 
vectors in $PT$ should be perpendicular to column vectors in $BT
$ (i.e., $PTT^TB^T \approx 0$).\\
\textbullet $\:$ The transformed embeddings preserve the distances between any two 
vectors in the matrix $A$.

Let $X = TT^T$, we can capture these two objectives as the following semi-positive definite programming problem.   
\begin{equation}
\label{eq:sdp1}
\min_{X\succeq 0}  \| AXA^T-AA^T \|_F^2 + \lambda \|PXB^T\|_F^2 \\
\end{equation} 
where $\|\dot \|_F$ is the Frobenius norm,  
the first term ensures that the pairwise distances are preserved, and the second term induces the biases to be small on the seed words.    
The user-specified parameter $\lambda$
balances the two terms.

Directly solving this SDP optimization problem
is challenging.   
In practice, the dimension of matrix $A$ is in the scale of 400,000 $\times$ 300.
The dimensions of the matrices $AXA^T$ and $AA^T$ are $400,000 \times 400,000$,
causing computational and memory issues.
We conduct singular value decomposition on $A$, such that $A=U\Sigma V^T$, where $U$ and $V$ are orthogonal matrices and $\Sigma$ is a diagonal matrix.
\begin{equation}
\label{eq:pairdist}
\begin{split}
\|AXA^T-AA^T\|_F^2 &= \|A(X-I)A^T\|_F^2\\ 
&= \|U\Sigma V^T (X-I) V \Sigma U^T\|_F^2 \\
&=  \|\Sigma V^T (X-I) V \Sigma\|_F^2.
\end{split}
\end{equation}
The last equality follows the fact that $U$ is an orthogonal matrix ($\|UXU^T\|_F^2 = tr(UXU^TUX^TU^T)=tr(UXX^TU^T) = tr(XX^TU^TU) = \|X\|_F^2$.)

Substituting Eq.  \eqref{eq:pairdist} to Eq. \eqref{eq:sdp1} gives 
\begin{equation}
\label{eq:sdp2}
\min_{X\succeq 0}   \|\Sigma V^T (X-I) V \Sigma\|_F^2 + \lambda \|PXB^T\|_F^2
\end{equation}
Here $\Sigma V^T (X-I) V \Sigma$ is a $300 \times 300 $ matrix and can be solved efficiently. The solution $T$ is the debiasing transformation of the word embedding.   

\paragraph{Experimental validation.} To validate our debiasing algorithm, we asked Turkers to suggest words that are likely to reflect gender stereotype (e.g. \emph{manager}, {\em nurse}). We collected 438 such words, of which a random setup of 350 are used for training as the columns of the $P$ matrix. The remaining are used for testing. Figure~2 illustrates the results of the algorithm. The blue circles are the 88 gender-stereotype words suggested by the Turkers which form our held-out test set. The green crosses are a random sample of background words that were not suggested to have stereotype. Most of the stereotype words lie close to the $y = 0$ line, consistent with them lies near the midpoint between {\em he} and {\em she}. In contrast the background points were substantially less affected by the debiasing transformation. 

We use variances to quantify this result. For each test word (either gender-stereotypical or background) we project it onto the {\em he} - {\em she} direction. Then we compute the variance of the projections in the original embedding and after the debiasing transformation. For the gender-stereotype test words, the variance in the original embedding is 0.02 and the variance after the transformation is 0.001. For the background words, the variance before and after the transformation was 0.005 and 0.0055 respectively. This demonstrates that the transformation was able to reduce gender stereotype.   
 
Lastly to verify that the debiasing transformation $T$ preserves the desirable geometric structure of the embedding, we tested the transformed embedding a several standard benchmarks that measure whether related words have similar embeddings  as well as how well the embedding performs in analogy tasks. Table 2 shows the results on the original and the transformed embeddings and the transformation does not negatively impact the performance.
\begin{figure}
\includegraphics[width=\linewidth]{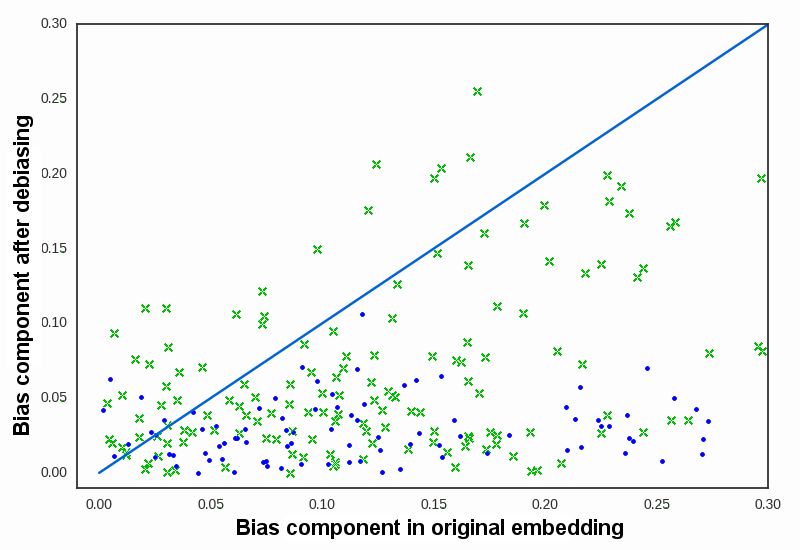}
	\caption{The changes of word vectors on the gender direction.  
The $x$ and $y$ axes show the absolute values of the projections onto the 
\emph{he-she} direction before and after debasing. 
The solid line is the diagonal. 
The blue `$\bullet$' are gender-stereotypical words in the 
test set, and the green `x' are randomly selected other words that were not suggested to be gender related.  
}
\label{fig:gnews_debias_bias}
\vspace{-0.5cm}
\end{figure}
\begin{table}
\vspace{-0.4cm}
\centering
\begin{tabular}{c|ccccc}
  Model&& RG & WS353 & RW & MSR-analogy  \\
  \hline
  Before && 0.761 & 0.700 & 0.471 & 0.712\\
  After &&0.764 & 0.700 & 0.472 & 0.712\\
\end{tabular}%
\caption{The columns show the performance of the 
	word embeddings on the standard evaluation metrics. RG \cite{rubenstein1965contextual}, RW \cite{luong2013better}, WS353 \cite{finkelstein2001placing}, MSR-analogy \cite{mikolov2013linguistic}\label{tab:variance}}
\end{table}

\clearpage

\bibliography{bibfile}
\bibliographystyle{icml2016}

\end{document}